# A Transformer-based representation-learning model with unified processing of multimodal input for clinical diagnostics

Hong-Yu Zhou[1,&], Yizhou Yu[1,&,*], Chengdi Wang[2,&,*], Shu Zhang[3], Yuanxu Gao[4,5,6], Jia Pan[1], Jun Shao[2], Guangming Lu[8], Kang Zhang[4,5,6,7*], and Weimin Li[2,*]

[1]Department of Computer Science, The University of Hong Kong, Pokfulam, Hong Kong, China
[2]Department of Respiratory and Critical Care Medicine, Med-X Center for Manufacturing, Frontiers Science Center for Disease-related Molecular Network, West China Hospital, Sichuan University, Chengdu, China, 610041
[3]AI Lab, Deepwise Healthcare, Beijing, China, 100080
[4]Zhuhai International Eye Center, Zhuhai People's Hospital and the First Affiliated Hospital of Faculty of Medicine, Macau University of Science and Technology and University Hospital, Guangdong, China, 000000
[5]Department of Big Data and Biomedical Artificial Intelligence, College of Future Technology, Peking University, Beijing, China, 100871
[6]Guangzhou Laboratory, Guangzhou, China, 510005
[7]Clinical Translational Research Center, West China Hospital, Sichuan University, Chengdu, China, 610041
[8]Department of Medical Imaging, Jinling Hospital, Nanjing University School of Medicine, Nanjing, Jiangsu, China, 210093

[&]These authors contributed equally
*Corresponding authors, weimin003@scu.edu.cn; kang.zhang@gmail.com; chengdi_wang@scu.edu.cn; yizhouy@acm.org

**During the diagnostic process, clinicians leverage multimodal information, such as chief complaints, medical images, and laboratory-test results. Deep-learning models for aiding diagnosis have yet to meet this requirement. Here we report a Transformer-based representation-learning model as a clinical diagnostic aid that processes multimodal input in a unified manner. Rather than learning modality-specific features, the model uses embedding layers to convert images and unstructured and structured text into visual tokens and text tokens, and bidirectional blocks with intramodal and intermodal attention to learn a holistic representation of radiographs, the unstructured chief complaint and clinical history, structured clinical information such as laboratory-test results and patient demographic information. The unified model outperformed an image-only model and non-unified multimodal diagnosis models in the identification of pulmonary diseases (by 12% and 9%, respectively) and in the prediction of adverse clinical outcomes in patients with COVID-19 (by 29% and 7%, respectively). Leveraging unified multimodal Transformer-based models may help streamline triage of patients and facilitate the clinical decision process.**

One-sentence editorial summary (to appear right below the title of your Article on the journal's website):
**A Transformer-based representation-learning model that processes multimodal input in a unified manner outperformed non-unified multimodal models in two clinical diagnostic tasks.**

It has been a common practice in modern medicine to utilize multimodal clinical information for medical diagnosis. For instance, apart from chest radiographs, thoracic physicians need to take into account each patient's demographics (e.g., age and gender), the chief complaint (e.g., history of present and past illness), and the laboratory-test report to make accurate diagnostic decisions. In practice, abnormal radiographic patterns are first associated with symptoms mentioned in the chief complaint or abnormal results in the laboratory-test report. Then, physicians rely on their rich domain knowledge and years of training to make optimal diagnoses by jointly interpreting such multimodal data [1,2]. The importance of exploiting multimodal clinical information has been extensively verified in the literature [3-10] in different specialties, including but not limited to, radiology, dermatology, and ophthalmology.

The above multimodal diagnostic workflow requires enormous expertise, which may not be available in geographic regions with limited medical resources. Meanwhile, simply increasing the workload of experienced physicians and radiologists would inevitably exhaust their energy and thus increase the risk of misdiagnosis. To meet the increasing demand for precision medicine, machine learning techniques [11] have become the de facto choice for automatic yet intelligent medical diagnosis. Among them, the unprecedented development of deep learning [12,13] endows machine learning models with the ability to detect diseases from medical images near or at the level of human experts [14-18].

Although AI-based medical image diagnosis has achieved tremendous progress in recent years, it is still debatable how to jointly interpret medical images and their associated clinical context. As illustrated in **Fig. 1a**, current multimodal clinical decision support systems [19-23] mostly lean upon a non-unified way to fuse information from multiple sources. Given a set of input data from different sources, these approaches first roughly divide them into three basic modalities, i.e., images, narrative text (e.g., the chief complaint that includes the history of present and past illness), and structured fields (e.g., demographics and laboratory-test results). Next, a text structuralization process is introduced to transform the narrative text into structured tokens. Then, data in different modalities are fed to different machine learning models to produce modality-specific features or predictions. Finally, a fusion module is employed to unify these modality-specific features or predictions for making final diagnostic decisions. In practice, according to whether joining multiple input modalities at the feature or prediction level, these non-unified methods can be further categorized into early [19-22] or late fusion [23] methods.

One glaring issue of early and late fusion methods is that they separate the multimodal diagnostic process into two relatively independent stages: modality-specific model training and diagnosis-oriented fusion. However, such a design has one obvious limitation: the inability to encode the connections and associations among different modalities. Another non-negligible drawback of these non-unified approaches lies in the text structuralization process, which is cumbersome and still labor-intensive, even with the assistance of modern natural language processing (NLP) tools. On the other hand, Transformer-based architectures [24] are poised to broadly reshape natural language processing [25] and computer vision [26]. Compared to convolutional neural networks [27] and word embedding algorithms [28,29], Transformers [24] impose few assumptions about the input data form and thus have the potential to learn higher-quality feature representations from multimodal input data. More importantly, the basic architectural component in Transformers (i.e., the self-attention block) remains nearly unchanged across different modalities [25,26], providing an opportunity to build a unified yet flexible model to conduct representation learning on multimodal clinical information.

In this paper, we present IRENE, a unified AI-based medical diagnostic model designed to make decisions by jointly learning holistic representations of medical images, unstructured chief complaint, and structured clinical information. To the best of our knowledge, IRENE is the first medical diagnostic approach that uses a single, unified AI model to conduct holistic representation learning on multimodal clinical information simultaneously, as shown in **Fig. 1a**. At the core of IRENE are the unified multimodal diagnostic Transformer (MDT) and bi-directional multimodal attention blocks. MDT is a new Transformer stack that directly produces diagnostic results from multimodal input data. This new algorithm enables IRENE to take a different approach from previous non-unified methods by learning holistic representations from multimodal clinical information progressively while eliminating separate paths for learning modality-specific features. In addition, MDT endows IRENE with the ability to perform representation learning on top of unstructured raw text, which avoids tedious text structuralization steps in non-unified approaches. For better handling the differences among modalities, IRENE introduces bi-directional multimodal attention to bridge the gap between token-level modality-specific features and high-level diagnosis-oriented holistic representations by explicitly encoding the interconnections among different modalities. This explicit encoding process can be regarded as a complement to the holistic multimodal representation learning process in MDT.

As shown in **Fig. 2a**, MDT is primarily composed of embedding layers, bi-directional multimodal blocks, and self-attention blocks. Because of MDT, IRENE has the ability to jointly interpret multimodal clinical information simultaneously. Specifically, a free-form embedding layer is employed to convert unstructured and structured texts into uniform text tokens (cf. **Fig. 2b**). Meanwhile, a similar tokenization procedure is also applied to each input image (cf. **Fig. 2c**). Next, two bi-directional multimodal blocks (cf. **Fig. 2d**) are stacked to learn fused mid-level representations across multiple modalities. In addition to computing intra-modal attention among tokens from the same modality, these blocks also explicitly compute inter-modal attention among tokens across different modalities (cf. **Fig. 2e**). These intra- and inter-modal attentional operations are consistent with daily clinical practices, where physicians need to discover interconnected information within the same modality as well as across different modalities. In reality, these connections are often hidden among local patterns, such as words in the chief complaint and image regions in radiographs, and different local patterns may refer to the same lesion or the same disease. Therefore, such connections provide mutual confirmations of clinical evidences and are helpful to both clinical and AI-based diagnosis. In bi-directional multimodal attention, each token can be regarded as the representation of a local pattern, and token-level intra- and inter-modal attention respectively capture the interconnections among local patterns from the same modality and across different modalities. In comparison, previous non-unified methods make diagnoses on top of separate global representations of input data in different modalities, and thus cannot exploit the underlying local interconnections. Finally, we stack ten self-attention blocks (cf. **Fig. 2f**) to learn multimodal representations.

IRENE shares some common traits with vision-language fusion models [29-33], both of which aim to learn a joint multimodal representation. However, one most noticeable difference exists in the roles of different modalities. IRENE is designed for the scenario where multiple modalities supply complementary semantic information, which can be fused and utilized to improve prediction performance. On the contrary, recent vision-language fusion approaches [31-33] heavily rely on the distillation and exploitation of common semantic information among different modalities to provide supervision for model training.

We validate the effectiveness of IRENE on two tasks (cf. **Fig. 1b**): a) pulmonary disease identification and b) adverse clinical outcome prediction of COVID-19 patients. In the first task, IRENE outperforms previous image-only and non-unified diagnostic counterparts by approximately 12% and 9% (cf. **Fig. 1c**), respectively. In the second task, we require IRENE to predict adverse clinical events of COVID-19 patients, i.e., admission to the intensive care unit (ICU), mechanical ventilation (MV) therapy, and death. Different from the first task, the second task relies more on textual clinical information. In this scenario, IRENE significantly outperforms non-unified approaches by over 7% (cf. **Fig. 1d**). Particularly noteworthy is the nearly 10-percent improvement that IRENE achieves on death prediction, which we believe will have an impact in assisting doctors to take immediate steps for saving COVID-19 patients. When compared to human experts (cf. **Fig. 1e**) in pulmonary disease identification, IRENE clearly surpasses junior physicians (with < 7 years of experience) in the diagnosis of all eight diseases while delivering a performance comparable to or better than that of senior physicians (with more than 7 years of experience) on six diseases.

**Results**
**Dataset characteristics for multimodal diagnosis.** The first dataset focuses on pulmonary diseases. We retrospectively collected consecutive chest X-rays from 51,511 patients between November 27, 2008, and May 31, 2019, at West China Hospital, which is the largest tertiary medical center in western China covering a 100 million population. Each patient is associated with at least one radiograph, a short piece of unstructured chief complaint, history of present and past illness, demographics, and a complete laboratory-test report. The dataset is built for eight pulmonary diseases, including chronic obstructive pulmonary disease (COPD), bronchiectasis, pneumothorax, pneumonia, interstitial lung disease (ILD), tuberculosis, lung cancer, and pleural effusion. Discharge diagnoses are extracted from discharge summary reports following the standard process described in previous study [16], and taken as the ground-truth disease labels. The discharge summary reports were produced as follows. An initial report was written by a junior physician, which was then reviewed and confirmed by a senior physician. In case of any disagreement, the final decision was made by a departmental committee comprised of at least three senior physicians.

The built dataset consists of 72,283 data samples, among which 40,126 samples are normal. The distribution of diseases (i.e., the number of relevant cases) is as follows: COPD (4,912), bronchiectasis (676), pneumothorax (2,538), pneumonia (21,409), ILD (3,283), tuberculosis (938), lung cancer (2,651) and pleural effusion (4,713). The performance metric is the area under the receiver operating characteristic curve (AUROC). We split this dataset into training, validation, and testing sets according to each patient's admission date. Specifically, the training set includes 44,628 patients admitted between November 27, 2008, and June 1, 2018. And the validation set includes 3,325 patients admitted between June 2, 2018 and December 01, 2018. Finally, the trained and validated IRENE system is tested on 3,558 patients admitted between December 02, 2018 and May 31, 2019. Although this is a retrospective study, our data splitting scheme follows the practice of a prospective study, thus creates a more challenging and realistic setting to verify the effectiveness of different multimodal medical diagnosis systems, in comparison to data splitting schemes based on random sampling.

The second dataset MMC (i.e., multimodal COVID-19 dataset) [19], on which IRENE is trained and evaluated, consists of chest CT images and structured clinical information (e.g., chief complaint that comprises comorbidities and symptoms, demographics, laboratory-test results, etc) collected from COVID-19 patients. The CT images are associated with inpatients with laboratory-confirmed COVID-19 infection between December 27, 2019 and March 31, 2020. There are three types of adverse events that could happen to patients in MMC, which are admission to ICU, mechanical ventilation (MV), and death. The training and validation sets came from 17 hospitals, and the training set has 1,164 labeled cases (70%) while the validation set has 498 labeled ones (30%). Next, we chose the trained model with the best performance on the validation set and test it on the independent testing set, which is comprised of 700 cases collected from 9 external medical centers. The distribution of the three events in the testing set is as follows: ICU (155), MV (94), Death (59). This is an imbalanced classification problem where the majority of patients does not have any adverse outcomes. Against this background, we use the area under the precision-recall curve (AUPRC) instead of AUROC as the performance metric, which focuses more on identifying adverse events (i.e., ICU, MV, and Death).

**Pulmonary disease identification. Table 1** and **Fig. 3** present the experimental results from IRENE and other methods on the dataset for pulmonary disease identification. As shown in **Table 1**, IRENE significantly outperforms the image-only model, traditional non-unified early [19] and late fusion [23] methods, and two recent state-of-the-art Transformer-based multimodal methods (i.e., Perceiver [30] and GIT [33]) in identifying pulmonary diseases. Generally speaking, IRENE achieved the highest mean AUROC (0.924, [95% CI: 0.921, 0.927]), about 12% higher than the image-only model (0.805, [95% CI: 0.802, 0.808]) that only takes radiographs as the input. In comparison to diagnostic decisions made by non-unified early fusion (0.835, [95% CI: 0.832, 0.839]) and late fusion (0.826, [95% CI: 0.823, 0.828]) methods, IRENE maintained an advantage of 9% at least. Comparing IRENE to GIT (0.848, [95% CI: 0.844, 0.850]), we observed an advantage of over 7%. Even when compared to Perceiver, the Transformer-based multimodal classification model developed by DeepMind, IRENE still delivered competitive results, surpassing Perceiver (0.858 [95% CI: 0.855, 0.861]) by over 6%. When carefully checking each disease and comparing IRENE against the previous best result among all five baselines, we observed that among all eight pulmonary diseases, IRENE achieved the largest improvements on bronchiectasis (12%), pneumothorax (10%), ILD (10%), and tuberculosis (9%).

We also compared IRENE against human experts, who were divided into two groups, one group of two junior physicians (with < 7 years of experience) and the second group of two senior physicians (with ⩾ 7 years of experience). For better comparison, we present the average performance within each group in **Fig. 1e**. Specifically, we extract annotations by human experts from electronic discharge diagnosis records. Note that all physicians from the reader study did not participate in data annotation. We see that IRENE exhibits advantages over the junior group on all eight pulmonary diseases, especially in the diagnosis of bronchiectasis (Junior, [FPR: 0.29, TPR: 0.58]), pneumonia (Junior, [FPR: 0.37, TPR: 0.76]), ILD (Junior, [FPR: 0.09, TPR: 0.63]), and pleural effusion (Junior, [FPR: 0.35, TPR: 0.86]), where FPR and TPR stand for the false and true positive rates, respectively. Compared to the senior group, IRENE is advantageous in the diagnosis of pneumonia (Senior, [FPR: 0.21, TPR: 0.80]), tuberculosis (Senior, [FPR: 0.07, TPR: 0.17]), and pleural effusion (Senior, [FPR: 0.25, TPR: 0.77]). In addition, IRENE performed comparably with senior physicians on COPD (Senior, [FPR: 0.07, TPR: 0.76]), ILD (Senior, [FPR: 0.09, TPR: 0.71]), and pneumothorax (Senior, [FPR: 0.08, TPR: 0.79]) while showing slightly worse performance on bronchiectasis (Senior, [FPR: 0.12, TPR: 0.82]) and lung cancer (Senior, [FPR: 0.08, TPR: 0.73]).

**Adverse clinical outcome prediction of COVID-19 patients.** Triage of COVID-19 patients heavily depends on joint interpretation of chest CT scans and other non-imaging clinical information. In this scenario, IRENE exhibited even more advantages than it did in the pulmonary disease identification task. As shown in **Table 2**, IRENE consistently achieved impressive performance improvements on the prediction of the three adverse clinical outcomes of COVID-19 patients, i.e., admission to ICU, mechanical ventilation, and death. In terms of mean AUPRC, IRENE (0.592, [95% CI: 0.500, 0.682]) outperformed the image-only model (0.307, [95% CI: 0.237, 0.391]), early fusion model [22] (0.521, [95% CI: 0.435, 0.614]), and late fusion model [23] (0.503, [95%: 0.422, 0.598]) by nearly 29%, 7%, and 9%, respectively. As for specific clinical outcomes, IRENE (0.712, [95% CI: 0.587, 0.834]) achieved about 5-percent AUPRC gain over the non-unified early fusion method (0.665, [95% CI: 0.548, 0.774]) in the prediction of admission to ICU. Similarly, in the prediction of MV, IRENE achieves an over 6-percent performance improvement when compared to the early fusion model. Last but not the least, IRENE (0.441, [95% CI: 0.270, 0.617]) is much more capable of predicting death than the image-only model (0.192, [95% CI: 0.073, 0.333]), early fusion model (0.346, [95%: 0.174, 0.544]), and late fusion model (0.335, [95% CI: 0.168, 0.554]). Compared to two Transformer-based multimodal models, i.e., GIT and Perceiver, we observe an advantage of over 6% on average.

**Impact of different modules and modalities in IRENE.** To investigate the impact of different modules and modalities, we conducted thorough ablative experiments and reported their results in **Table 3**. First of all, we investigated the impact of bi-directional multimodal attention blocks (rows 0-2). We found that replacing all bi-directional multimodal attention blocks with self-attention blocks led to about 7-percent performance drop (from 0.924 to 0.858) in pulmonary disease identification. This phenomenon verified our intuition that directly learning progressively fused representations from raw data would deteriorate the diagnosis performance. On the contrary, simply increasing the number of bi-directional multimodal attention blocks from two to six did not bring obvious performance improvements (from 0.924 to 0.905), indicating that using two successive bi-directional multimodal attention blocks can be an optimal choice in IRENE. In row 3, we present the result of using uni-directional attention (i.e., text-to-image attention). Comparing row 0 with row 3, we see that our bi-directional design brings a 4-percent performance gain (from 0.884 to 0.924). Next, we studied the impact of clinical texts (rows 4 and 5). The first observation was that utilizing the complementary narrative chief complaint substantially boosts the diagnostic performance because removing chief complaint from the input data reduced model performance by 6% (from 0.924 to 0.860). Apart from chief complaint, we also studied the impact of laboratory-test results (row 5). We observed that including laboratory-test results brings about a 4-percent performance gain (from 0.882 to 0.924). Then, we investigated the impact of tokenization procedures. We saw that modelling the chief complaint and laboratory-test results of a patient as a sequence of tokens (row 0) did perform better than directly passing an averaged representation (row 6) to the model. This improvement brought by the tokenization of chief complaint and laboratory-test results verified the advantage of token-level intra- and inter-modal bi-directional multimodal attention, which exploited local interconnections among the word tokens of the clinical text and the image patch tokens of the radiograph in the input data. In the end, we investigate the impact of the input image in IRENE (row 7) and observe a dramatic performance drop (from 0.924 to 0.543). This phenomenon indicated the vital role of the input radiograph in pulmonary disease identification. We then investigated the impact of chief complaints and laboratory-test results on each respiratory disease (cf. Extended Data Fig. 1). When we removed either chief complaints or the laboratory-test results from the input, the performance decreased on each disease. Specifically, we found introducing the chief complaint can be most helpful to the diagnosis of pneumothorax, lung cancer, and pleural effusion, while the laboratory-test results affect the diagnosis of bronchiectasis and tuberculosis the most. Clinical interpretations can be found in Supplementary Note 1.

**Attention visualization results. Fig. 3** provides attention visualization results for a case with COPD. From **Fig. 3a**, we see that the image modality (i.e., the radiograph) plays a significant role in the diagnostic process, and its weight is nearly 80% in the final decision. Besides, the chief complaint is the second most important factor, accounting for roughly 16% weight. As **Fig. 3b** shows, PaO2 (i.e., oxygen pressure in arterial blood) and PaCO2 (i.e., partial pressure of carbon dioxide in arterial blood) are the two most important laboratory-test items, which are consistent with the observations reported in the literature [34]. Nonetheless, we see that the total weight of the remaining 90 test items is quite large, whose distribution over these 90 laboratory-test items is nearly uniform. The reason may be that these laboratory-test items help rule out other diseases. **Fig. 3c** shows that from the perspective of IRENE, age is a more critical factor than sex. **Fig. 3d** provides the attention map of the radiograph. We see that IRENE refers to hilar enlargement, hyper-expansion, and flattened diaphragm as the most important evidences for the diagnosis of COPD. Besides, IRENE also identifies large black areas due to bullae as relatively important evidence. **Fig. 3e** summarizes the experimental results with and without cross attention, where we present the sum of similarity scores of important (top 25%) tokens (i.e., words and image patches) with the CLS token. We found that with cross attention, the sum of similarity scores becomes larger, indicating that cross attention improves the identification of important tokens compared to the model without cross attention. **In Fig. 3f**, IRENE recognizes "sputum", "dyspnea", and "years" as the three most important words in the chief complaint. **Fig. 3g** provides the cross-attention maps between each of the top three important words and the image. As the chief complaint and radiograph provide complementary information, keywords in the chief complaint, such as "sputum", "dyspnea" and "years", may not be directly observable in radiographs. As a result, it is infeasible to use cross-attention maps to identify image regions semantically corresponding to such keywords. Nonetheless, cross-attention maps can offer clinical insights to a certain extent. For example, in Figure 3g, the word "sputum" is primarily associated with the trachea and the lower pulmonary lobes in the image. The high attention area of the trachea is reasonable because trachea is often the location where sputum occurs. The high attention region in the left lower lobe has reduced vascular markings, while both the left and right lower lobes of the lungs are hyperinflated. Hyperinflated lungs and reduced vascular markings are common symptoms of COPD, which often has abnormal sputum production. Our model has associated the word "dyspnea" with most areas of the lungs in the image because dyspnea can be caused by a variety of pulmonary abnormalities that could occur anywhere in the lungs. Our model has also identified the areas surrounding the bronchi as the image regions associated with the word "years", which implies "years" should be associated with chronic diseases, such as chronic bronchitis, which is often part of COPD.

**Discussion**
**IRENE is more effective than the previous non-unified early and late fusion paradigm in multimodal medical diagnosis.** This is the most prominent observation obtained from our experimental results, and it holds in both tasks of pulmonary disease identification and triage of COVID-19 patients. Specifically, IRENE

outperforms previous early fusion and late fusion methods by an average of 9% and 10%, respectively, for identifying pulmonary diseases. Meanwhile, IRENE achieves about 3-percent performance gains on all eight diseases, and substantially improves the diagnostic performance on four diseases (i.e., bronchiectasis, pneumothorax, ILD, and tuberculosis) by boosting their AUROC by over 10%. We believe these prominent performance benefits are closely related to several capabilities of IRENE. First, IRENE is built on top of a unified Transformer, i.e., MDT. MDT directly produces diagnostic decisions from multimodal input data, and learns holistic multimodal representations progressively and implicitly. In contrast, the traditional non-unified approach decomposes the diagnosis problem into several components, which, in most cases, consist of data structuralization, modality-specific model training, and diagnosis-oriented fusion. In practice, these components are hard to optimize and may prevent the model from learning holistic and diagnosis-oriented features. Second, inspired by physicians' daily activities, IRENE applies intra- and bi-directional inter-modal attention to tokenized multimodal data for exploiting the local interconnections among complementary modalities. On the contrary, the previous non-unified paradigm directly makes use of the extracted global modality-specific representations or predictions for diagnosis. In practice, the token-level attentional operations in proposed bi-directional multimodal attention help capture and encode the interconnections among the local patterns of different modalities into the fused representations. Last but not the least, IRENE is designed to conduct representation learning directly on unstructured raw texts. In contrast, the previous non-unified approach relies on non-clinically pre-trained NLP models to provide word embeddings, which inevitably distracts the diagnosis system from its intended functionality.

The superiority of the aforementioned abilities has been partly verified in the second task, i.e., adverse clinical outcome prediction of COVID-19 patients. From **Table 2**, we see that IRENE holds a 7-percent average performance gain over the early fusion approach and an average of 9-percent advantage over the late fusion one. This performance gain is a little lower than that in the pulmonary disease identification task as there are no unstructured texts in the MMC dataset that IRENE can utilize. Nonetheless, IRENE can still leverage its unified and bi-directional multimodal attention mechanisms to better serve the goal of rapid triage of COVID-19 patients. For example, IRENE boosts the performance of MV and death prediction by 7% and 10%, respectively. Such substantial performance improvements brought by IRENE are valuable in the real world for allocating appropriate medical resources to patients in a timely manner, as medical resources are usually limited in the COVID-19 pandemic.

**IRENE provides a better Transformer-based choice for jointly interpreting multimodal clinical information.** We compare IRENE to GIT [33] and Perceiver [30], two representative Transformer-based models that fuse multimodal information for classification. GIT performs multimodal pre-training on tens of millions of image-text pairs by utilizing the common semantic information among different modalities as supervision signals. However, these characteristics have two obvious deficiencies in the medical diagnosis scenario. First, it is much harder to access multimodal medical data in the amount of the same order of magnitude. Second, multimodal data in the medical diagnosis scenario provide complementary instead of common semantic information. Thus, it is impractical to perform large-scale multimodal pre-training, as in GIT, using a limited amount of medical data. These deficiencies are also reflected in the experimental results. For instance, the average performance of GIT is about 7- and 8-percent lower than IRENE in the pulmonary disease identification task and adverse outcome prediction of COVID-19 task, respectively. These advantages show that token-level bi-direction multimodal attention in IRENE can effectively utilize limited amount of multimodal medical data and exploit complementary semantic information.

Perceiver simply concatenates multimodal input data and takes the resulting 1D sequence as the input instead of learning fused representations among modality-specific low-level embeddings as in IRENE. This poses a potential problem: the modality that makes up the majority of the input would have a larger impact on final diagnostic results. For example, since an image often has a much larger number of tokens than a text, Perceiver would inevitably assign more weight to the image instead of the text when making predictions. However, it is not always true that images play a more important role in daily clinical decisions. To some extent, this point is also reflected in our experimental observations. For example, Perceiver yields clear performance improvements (2-percent gain on average in **Table 1**) over the early fusion model in identifying pulmonary diseases whereas the input radiograph serves as the main information source. But in the task of rapid triage of COVID-19 patients, the performance of Perceiver is only comparable to that of the early fusion method. The underlying reason is that CT images are not as helpful in this task as radiographs in pulmonary disease identification. In contrast, IRENE demonstrates satisfactory performance in both tasks by learning holistic multimodal representations through bi-directional multimodal attention. Our method encourages features from different modalities to evenly blend into each other, which prevents the learned representations from being dominated by high-dimensional inputs.

**IRENE helps reduce the reliance on text structuralization in the traditional workflow.** In traditional non-unified multimodal medical diagnosis methods, the usual way to deal with unstructured texts is text structuralization. Recent text structuralization pipelines in non-unified approaches [19-23] severely rely on artificial rules and the assistance of modern NLP tools. For example, text structuralization requires human annotators to manually define a list of alternate spellings, synonyms, and abbreviations for structured labels. On top of these preparations, specialized NLP tools are developed and applied to extract structured fields from unstructured texts. As a result, text structuralization steps are not only cumbersome but also costly in terms of labor and time. In comparison, IRENE abandons such tedious structuralization steps by directly accepting unstructured clinical texts as part of the input.

**Outlook**
In conclusion, although NLP technologies particularly Transformers have contributed significantly to latest AI diagnostic tools using either text-based electronic health records [35] or images [36], this study describes an AI framework consisting of a unified multimodal diagnostic Transformer (MDT) and bi-directional multimodal attention blocks. This new algorithm enables IRENE to take a different approach from previous non-unified methods by progressively learning holistic representations for multimodal clinical data while eliminating separate paths for learning modality-specific features in non-unified techniques.

In real-world scenarios, IRENE may help streamline patient care, such as triaging patients and differentiating between those patients who are likely to have a common cold from those who need urgent intervention for a more severe condition. Furthermore, as the algorithms become increasingly refined, these frameworks could become a diagnostic aid for physicians and assist in cases of diagnostic uncertainty or complexity, thus not only mimicking physician reasoning but further enhancing it. The impact of our work may be most obvious in areas where there are few and uneven distributions of healthcare providers relative to the population.

In the following, we point out several limitations that need to be considered during the deployment of IRENE in clinical workflows and provide some insights to address them. First, currently used datasets are limited in both size and diversity. To resolve this issue, we may have to collect more data from additional medical institutions, medical devices, countries, and ethnic groups, with which we can train IRENE to enhance its generalization ability under a broader range of clinical settings. Second, the clinical benefits of IRENE need to be further verified. Thus multi-institutional multi-national studies can further validate the clinical utility of IRENE in real-world scenarios. Third, it is important to make IRENE adapt to a changing environment, such as dealing with rapidly mutating SARS-CoV-2 viruses. To tackle this challenge, we can train the model on multiple cohorts jointly or resort to other machine learning technologies, such as online learning. Last but not the least, IRENE fails to consider the problem of modal deficiency, where one or more modalities may be unavailable. To deal with this problem, we can refer to masked modeling [25]. For instance, during the training stage, we can randomly mask some modalities to imitate the absence of these modalities in clinical workflows.

**Methods**
**Image and textual clinical data.** In the pulmonary disease identification task, CXR images were collected from West China Hospital. All CXRs were collected as part of the patients' routine clinical care. For the analysis of CXR images, all radiographs were first de-identified to remove any patient-related information. The CXR images consisted of both an anterior-posterior view of CXR images. There are three types of textual clinical data: the unstructured chief complaint (i.e., history of present and past illness), demographics (age and gender), and laboratory-test results. Specifically, the chief complaint is unstructured while demographics and laboratory-test results are structured. We set the maximum length of the chief complaint to 40. If a patient's chief complaint has more than 40 words, we only take the first 40; otherwise, zero padding is used to satisfy the length requirement. There are 92 results in each patient's laboratory-test report (refer to Supplementary Note 2), most of which come from the blood test. We normalize every test result through min-max scaling so that every normalized value lies in [0, 1], where the minimum and maximum values in min-max scaling are determined using the training set. In particular, -1 denotes missing values.

In the second task, i.e., adverse clinical outcome prediction of COVID-19 patients, the available clinical data can be divided into four categories: demographics (age and gender), the structured chief complaint that consists of comorbidities (7) and symptoms (9), and laboratory-test results (19). Please refer to Supplementary Note 3 for more details. Following, we apply median imputation to fill in missing values.

Institutional Review Board (IRB)/Ethics Committees approvals were obtained from West China Hospital and all participating hospitals. All patients signed a consent form. The research was conducted in a manner compliant with the United States Health Insurance Portability and Accountability Act (HIPAA). It was adherent to the tenets of the Declaration of Helsinki and in compliance with the Chinese CDC policy on reportable infectious diseases and the Chinese Health and Quarantine Law.

**Baseline models.** We include five baseline models in our experimental performance comparisons, including the diagnosis model purely based on medical images (denoted as Image-only), the traditional non-unified early and late fusion methods with multimodal input data, and two recent state-of-the-art Transformer-based multimodal classification methods (i.e., GIT and Perceiver). The implementation details of them are as follows:

- **Image-only.** In the pulmonary disease identification task, we build the pure medical image based diagnosis model on top of ViT [26], one of the most well-known and widely adopted Transformer-based deep neural networks for image understanding. Our ViT-like network architecture has 12 blocks and each block consists of one self-attention layer [24], one multi-layer perceptron (MLP), and two layer normalization layers [37]. There are two fully-connected (FC) layers in each MLP, where the number of hidden nodes is 3,072. The input size of the first FC layer is 768. Between the two FC layers, we insert a GeLU activation function [38]. After each FC layer, we add a dropout layer [39], where we set the dropout rate to 0.3. The output size of the second FC layer is also 768. Each input image is divided into a number of 16×16 patches. The output CLS token is used for performing the final classification. We use the binary cross-entropy loss as the cost function during the training stage. Note that before the training stage, we perform supervised ViT pre-training on MIMIC-CXR [40] to obtain visual representations with more generalization power. In the task of rapid triage of COVID-19 patients, as in [22], we first segment pneumonia lesions from CT scans, then train multiple machine learning models (i.e., logistic regression, random forest, support vector machine, MLP, and LightGBM) using image features extracted from the segmented lesion areas, and finally choose the optimal model according to their performance on the validation set.

- **Non-unified early and late fusion.** There are a number of existing methods using the archetypical non-unified approach to fuse multimodal input data for diagnosis. For better adaptation to different scenarios, we adopt different non-unified models in different tasks. Specifically, we modified the early fusion method reported in previous study [19] for our first task (i.e., pulmonary disease identification). In practice, a ViT model extracts image features from radiographs, and the feature vector at its CLS token is taken as the representation of the input image. Similar to the image-only baseline, supervised pre-training on MIMIC-CXR [40] was applied to the ViT to obtain more powerful visual features before we carry out the formal task. To process the three types of clinical data (i.e., the chief complaint, demographics, and laboratory-test results), we employ three independent MLPs to convert different types of textual clinical data to features, which are then concatenated with the image representation. The rationale behind is that both images and textual data should be represented in the same feature space for the purpose of cross reference. Since the chief complaint includes unstructured texts, we first need to transform them into structured items. To achieve this goal, we train an entity recognition model to highlight relevant clinical symptoms in the chief complaint. Next, we use BERT [25] to extract features for all such symptoms, to which average pooling is applied to produce a holistic representation for each patient's chief complaint. Then, we use a three-layer MLP to further transform this holistic feature into a latent space similar to that of the image representation. The input size of this three-layer MLP is 768, and the output size is 512. The number of hidden nodes is 1,024. After each FC layer, we add a ReLU activation and a dropout layer with the dropout rate set to 0.3. Likewise, for laboratory-test results, we also apply an MLP with the same architecture but independent weight parameters to transform those test results into a one-dimensional feature vector. The input size of this laboratory-test MLP is 92 and the output size is 512. The MLP model for demographics has two FC layers, where the input size is 2 and the output size is 512. The hidden layer has 512 nodes. The feature fusion module includes the concatenation operation and a three-layer MLP with the number of hidden nodes set to 1,024. The output from the MLP in the feature fusion module is passed to the final classification layer for making diagnostic decisions. During the training stage, we jointly train the ViT-like model and all MLPs using the binary cross-entropy loss. As for the late fusion baseline, we ensemble the predictions of the image- and text-based classifiers inspired by [23]. Specifically, we train a ViT model with radiographs and their associated labels. To construct the input to the text-based classifier, we concatenate laboratory-test results, demographics, and the holistic representation (obtained via averaging extracted features of symptoms, similar to the early fusion method) of the chief complaint. Then, we forward the constructed input through a three-layer MLP, whose input and output dimensions are 862 and 8, respectively. Then, we train the MLP with the same labels used for training the ViT model. Finally, we average the predicted probabilities of the image- and text-based classifiers to obtain the final prediction.

In the second task, we follow the early fusion method proposed in [22], where image features, structured chief complaint (comorbidities and symptoms), and laboratory-test results are concatenated as the input. Then, we train multiple machine learning models and choose the optimal model using those artificial rules introduced in [22]. For the late fusion baseline, we train five machine learning models (i.e., logistic regression,

- random forest, support vector machine, MLP, and LightGBM) following the protocol used in [22] for image features, structured chief complaints, and laboratory-test results, respectively. Then, we take the average of the predicted probabilities of these fifteen machine learning models as the adverse outcome prediction.
- **GIT.** GIT [33] is a generative image-to-text Transformer that unifies vision-language tasks. We take GIT-Base as a baseline in our comparisons. Its image encoder is a ViT-like Transformer, and its text decoder consists of six standard Transformer blocks [24]. In practice, we fine-tune the officially released pre-trained model on our own datasets. For fairness, we adopt the same set of fine-tuning hyper-parameters used for IRENE. In the pulmonary disease identification task, we first forward each radiograph through the image encoder to extract an image feature. Next, we concatenate this image feature with the averaged word embedding (using BERT) of the chief complaint as well as the feature vectors of the demographics and laboratory-test results. The concatenated features are then passed to the text decoder to make diagnostic predictions. In the task of adverse clinical outcome prediction of COVID-19 patients, we first average the image features of CT slices. Then, the averaged image feature is concatenated with the feature vectors of the clinical comorbidities and symptoms, laboratory-test results, and demographics. Next, we forward the concatenated multimodal features through the text decoder to predict adverse outcomes of patients with COVID-19.
- **Perceiver.** This is a very recent state-of-the-art Transformer-based model [30] from DeepMind, proposed for tackling the classification problem with multimodal input data. There also exists a variant of Perceiver [30], i.e., Perceiver IO [41], which introduces the output query on top of Perceiver to handle additional types of tasks. As making diagnostic decisions can be considered as a type of classification, we adopt Perceiver instead of Perceiver IO as one of our baseline models. Our Perceiver architecture follows the setting for ImageNet classification [42] in previous study [30], and has six cross-attention modules. Each cross-attention module is followed by a latent Transformer with six self-attention blocks. The input of Perceiver consists of two arrays: the latent array and byte array. Following [30], we initialize the latent array using a truncated zero-mean normal distribution with standard deviation set to 0.02 and truncation bounds set to [-2, 2]. The byte array consists of multimodal data. In the pulmonary disease identification task, we first flatten the input image into a one-dimensional vector. Then, we concatenate it with the averaged word embedding (using BERT) of the chief complaint as well as one-dimensional feature vectors of the input demographics and laboratory-test results. This results in a long one-dimensional vector, which is taken as the byte array. In the task of adverse clinical outcome prediction of COVID-19, we also flatten the input image into a one-dimensional vector, which is then concatenated with the feature vectors of the clinical comorbidities and symptoms, laboratory-test results, and demographics. The learning process of Perceiver can be summarized as follows: the latent array evolves by iteratively extracting higher-quality features from the input byte array by alternating cross-attention and latent self-attention computations. Finally, the transformed latent array serves as the representation used for diagnosis. Note that similar to the image-only and non-unified baselines, we pre-trained Perceiver on MIMIC-CXR [40]. During pre-training, we used zero padding in the byte array for the non-existent clinical text in every multimodal input.

**IRENE.** In practice, we forward multimodal input data (i.e., medical images and textual clinical information) to the MDT for acquiring prediction logits. During the training stage, we compute the binary cross-entropy loss between the logits and ground-truth labels. Specifically, we use pulmonary disease annotations (eight diseases) and real adverse clinical outcomes (3 clinical events) as the ground-truth labels in the first and second tasks, respectively.

MDT is a unified Transformer, which primarily consists of two starting layers for embedding the tokens from the input image and text, respectively, two stacked bi-directional multimodal attention blocks for learning fused mid-level representations by capturing interconnections among tokens from the same modality and across different modalities, ten stacked self-attention blocks for learning holistic multimodal representations and enhancing their discriminative power, and one classification head for producing prediction logits.

The multimodal input data in the pulmonary disease identification task (i.e. the first task) consist of five parts: a radiograph, the unstructured chief complaint that includes history of present and past illness, laboratory-test results, each patient's gender, and age, which are denoted as $x^I$, $x^{cc}$, $x^{lab}$, $x^{sex}$, and $x^{age}$, respectively. We pass $x^I$ to a convolutional layer, which produces a sequence of visual tokens. Next, we add standard learnable 1D positional embedding [21,23] and dropout to every visual token to obtain a sequence of image patch tokens $X^I_{1:N}$. Meanwhile, we apply word tokenization to $x^{cc}$ to encode each word from the unstructured chief complaint. Specifically, we use a pre-trained BERT [23] to generate an embedded feature vector for each word in $x^{cc}$, after which we obtain a sequence of word tokens $X^{cc}_{1:N^{cc}}$. We also apply a similar tokenization procedure to $x^{lab}$, where min-max scaling is first employed to normalize every component of $x^{lab}$. We then pass each normalized component to a shared linear projection layer to obtain a sequence of latent embeddings $X^{lab}_{1:N^{lab}}$. We also perform linear projections on $x^{sex}$ and $x^{age}$ to obtain encoded feature vectors $X^{sex}$ and $X^{age}$. Subsequently, we concatenate $\{X^{cc}_{1:N^{cc}}, X^{lab}_{1:N^{lab}}, X^{sex}, X^{age}\}$ together to produce a sequence of clinical text tokens $X^T_{1:\widehat{N}}$, where $\widehat{N} = N^{cc} + N^{lab} + 2$. In practice, we set $N^{cc}$ and $N^{lab}$ to 40 and 92, respectively.

As for the task of adverse clinical outcome prediction of COVID-19 patients, its multimodal input data also consist of five parts: a set of CT slices, structured chief complaint (comorbidities and symptoms), laboratory-test results, each patient's gender and age, which are denoted as $x^I$, $x^{cc}$, $x^{lab}$, $x^{sex}$, and $x^{age}$. Each CT slice is converted to a sequence of image patch tokens $X^I_{1:N}$ as in the first task. Different from the first task, the chief complaint is structured. To convert $x^{cc}$ to tokens, we conduct a shared linear projection to each component, which generates a sequence of embeddings $X^{cc}_{1:N^{cc}}$. A linear projection layer is applied to $x^{lab}$ to acquire $X^{lab}_{1:N^{lab}}$. As for $x^{sex}$ and $x^{age}$, we perform linear projections to obtain encoded $X^{sex}$ and $X^{age}$ as in the first task. Finally, we directly concatenate $\{X^{cc}_{1:N^{cc}}, X^{lab}_{1:N^{lab}}, X^{sex}, X^{age}\}$ to produce $\widehat{N}$ clinical text tokens $X^T_{1:\widehat{N}}$, where $\widehat{N} = N^{cc} + N^{lab} + 2$. We set $N^{cc}$ and $N^{lab}$ to 16 and 19, respectively.

The first two layers of MDT are two stacked bi-directional multimodal attention blocks. Suppose the input of the first bi-directional multimodal attention block consists of $X^l_I$ and $X^l_T$, where l (= 0) stands for the layer index, $X^0_I = X^I_{1:N}$ denotes the assembly of image patch tokens, and $X^0_T = X^T_{1:\widehat{N}}$ represents the bag of clinical text tokens. The process of generating the query, key, and value matrices for each modality in the bi-directional multimodal attention block is as follows:

$$Q^l_I, K^l_I, V^l_I = \text{LP}\left(\text{Norm}(X^l_I)\right),$$

$$Q^l_T, K^l_T, V^l_T = \text{LP}\left(\text{Norm}(X^l_T)\right),$$

where LP($\cdot$) and Norm($\cdot$) represent linear projection and layer normalization, respectively. The forward pass inside a bi-directional multimodal attention block can be summarized as:

$$\mathfrak{X}^l_I = \text{Attention}(Q^l_I, K^l_I, V^l_I) + \lambda\, \text{Attention}(Q^l_I, K^l_T, V^l_T),$$

$$\mathfrak{X}^l_T = \text{Attention}(Q^l_T, K^l_T, V^l_T) + \lambda\, \text{Attention}(Q^l_T, K^l_I, V^l_I),$$

where Attention($Q^l_I, K^l_I, V^l_I$) and Attention($Q^l_T, K^l_T, V^l_T$) capture the intra-modal connections in the image and text modalities, respectively. Attention($Q^l_I, K^l_T, V^l_T$) and Attention($Q^l_T, K^l_I, V^l_I$) dig out the inter-modal connections between the image and text. Next, both intra- and inter-modal connections are encoded into latent representations $\mathfrak{X}^l_I$ and $\mathfrak{X}^l_T$. We set $\lambda$ to 1.0 as it gave rise to the best performance in our preliminary experiments. Attention($Q, K, V$) includes two matrix multiplications and one scaled softmax operation:

$$\text{Attention}(Q, K, V) = \text{softmax}(\frac{QK^\mathsf{T}}{\sqrt{d_k}}V),$$

where T stands for the matrix transpose operator, $d_k$ is a scaling hyper-parameter, which is set to 64. Next, we introduce residual learning [43] and forward the resulting $\mathfrak{X}^l_I, \mathfrak{X}^l_T$ to the following normalization layer and MLP:

$$X^{l+1}_I = \text{MLP}\left(\text{Norm}(\mathfrak{X}^l_I)\right) + + X^l_I,$$

$$X^{l+1}_T = \text{MLP}\left(\text{Norm}(\mathfrak{X}^l_T)\right) + + X^l_T,$$

$X^{l+1}_I$ and $X^{l+1}_T$ are passed to the next bi-directional multimodal attention block as the input, resulting in $X^{l+2}_I$ and $X^{l+2}_T$. Then, we combine tokens in $X^{l+2}_I$ and $X^{l+2}_T$ to produce a bag of unified tokens, which are passed to the following self-attention blocks [24]. We also allocate multiple heads [24] in both bi-directional multimodal attention and self-attention blocks, where the number of heads is set to 12. This multi-head mechanism allows the model to perform attention operations in multiple representation subspaces simultaneously and aggregate the results afterwards.

At the end, we apply average pooling to the unified tokens generated from the last self-attention block to obtain a holistic multimodal representation for medical diagnosis. This representation is passed to a two-layer MLP to produce final prediction logits. During the training stage, we calculate the binary cross-entropy loss between these logits and their corresponding pulmonary disease annotations (the first task) or real adverse clinical outcomes (the second task). A loss function value is computed for every patient case. Specifically, in the first task, each patient case contains one radiograph and related textual clinical information. In the second task, each patient case involves multiple CT slices, and these CT slices share the same textual clinical information. We forward each CT slice and its accompanying textual clinical information to MDT to obtain one holistic representation. Since we have multiple CT slices, we obtain a number of holistic representations (equal to the number of CT slices) for the same patient. Then, we perform average pooling over these holistic representations to compute an averaged representation, which is finally passed to a two-layer MLP and the binary cross-entropy loss.

**Implementation details.** For the pulmonary disease identification task, we first resize each radiograph to 256×256 pixels during the training stage, then crop a random portion of each image, where the area ratio between the cropped patch and the original radiograph is randomly determined between 0.09 and 1.0. The cropped patch is resized to 224×224, after which a random horizontal flip is applied to increase the diversity of training data. In the validation and testing stages, each radiograph is first resized to 256×256 pixels, and then a square patch at the image center is cropped. The size of the square crop is 224×224. The processed radiographs are finally passed to the Image-only model, Non-unified-Chest, Perceiver, and IRENE as input images. In the task of adverse clinical outcome prediction of COVID-19 patients, the input images are CT scans. We first use the lesion detection and segmentation methodologies proposed in [44]. This is a deep learning algorithm based on a multi-view feature pyramid convolutional neural network [45,46], which performs lesion detection, segmentation, and localization. This neural network was trained and validated on 14,435 participants with chest CT images and definite pathogen diagnosis. On a per-patient basis, the algorithm showed superior sensitivity of 1.00 [95% CI: 0.95, 1.00] and an F1-score of 0.97 in detecting lesions from CT images of COVID-19 pneumonia patients. Adverse clinical outcomes of COVID-19 are presumed to be closely related to the characteristics of pneumonia lesion areas. For each patient's case, we crop a 3D CT subvolume by computing the minimum 3D bounding box enclosing all pneumonia lesions. Next, we resize all 3D subvolumes from different patients to a uniform size, which is 224×224×64. At the end, we sample 16 evenly spaced slices from every 3D subvolume along its third dimension.

Before we perform the formal training procedure, we pre-trained our MDT on MIMIC-CXR [40], as what we have done for the baseline models. Similar to Perceiver, during pre-training, we used zero padding for non-existent textual clinical information in every multimodal input. In the formal training stage, we use AdamW [47] as the default optimizer as we found empirically it gives rise to better performance on baseline models and IRENE. The initial learning rate is set to 3e-5 and the weight decay is 1e-2. We train each model for 30 epochs and decrease the initial learning rate by a factor of 10 at the 20-th epoch. The batch size is set to 256 in the training stage of both tasks. It is worth noting that in the task of adverse clinical outcome prediction of COVID-19, we first extract holistic feature representations from 16 CT slices (cropped and sampled from the same CT volume). Next, we apply average pooling to these 16 holistic features to obtain an averaged representation, which represents all pneumonia lesion areas in the entire CT volume. The binary cross-entropy loss is then computed on top of this averaged representation. During the training stage, we evaluate the model performance on the validation set and calculate the validation loss after each epoch. The model checkpoint that produces the lowest validation loss is saved and then tested on the testing set. We employ learnable positional embeddings in all ViT models. IRENE is implemented using PyTorch [48] and the training stage is accelerated using NVIDIA Apex with the mixed-precision strategy [49]. In practice, we can finish the training stage of either task within one day using four NVIDIA GPUs.

We adopted the standard attention analysis strategy for vision Transformers. For each layer in the Transformer, we average the attention weights across multiple heads (as we used multi-head self-attention in IRENE) to obtain an attention matrix. To account for residual connections, we add an identity matrix to each attention matrix and normalize the resulting weight matrices. Next, we recursively multiply the weight matrices from different layers of the Transformer. Finally, we obtain an attention map that includes the similarity between every input token and the CLS token. Since the CLS token is used for diagnostic predictions, these similarities indicate the relevance between the input tokens and prediction results, which can be used for visualization. For cross-attention results, we perform visualization with Grad-CAM [50].

Non-parametric bootstrap sampling is used to calculate 95% confidence intervals. Specifically, we repeatedly draw 1,000 bootstrap samples from the unseen test set. Each bootstrap sample is obtained through random sampling with replacement, and its size is the same as the size of the test set. We then compute AUROC (the first task) or AUPRC (the second task) on each bootstrap sample, after which we have 1,000 AUROC or AUPRC values. Finally, we sort these performance results and report the values at 2.5 and 97.5 percentiles, respectively.

To demonstrate the statistical significance of our experimental results, we first repeat the experiments of IRENE and the best performing baseline (i.e., Perceiver) five times with different random seeds. Then, we calculate P-values between the mean performance of IRENE and the best baseline results using the independent two-sample t-test (two-sided).

**Reporting Summary**. Further information on research design is available in the Nature Research Reporting Summary linked to this article.

**Data availability**
Restrictions apply to the availability of the developmental and validation datasets, which were used with permission of the participants for the current study. De-identified data may be available for research purposes from the corresponding authors on reasonable request.

**Code availability**
Code is available at https://github.com/RL4M/IRENE.

**Acknowledgements**
W.L. and C.W disclose support for the publication of this study from the National Natural Science Foundation of China (82100119, 92159302, 91859203), Y.Y. discloses support for the publication of this study from Hong Kong Research Grants Council through General Research Fund (Grant 17207722), K.Z. discloses support for the research described in this study from Macau Science and Technology Development Fund, Macao (0007/2020/AFJ, 0070/2020/A2, 0003/2021/AKP).

**Author contributions**
H.Z., Y.Y., K.Z., and W.L. conceived the idea and designed the experiments. H.Z., C.W., and S.Z. implemented and performed the experiments. H.Z., Y.Y., C.W., S.Z., J.P., J.S., Y.G., G.L., K.Z., and W.L. analyzed the data and experimental results. H.Z., Y.Y., C.W., Y.G., K.Z., and W.L. wrote the manuscript. All authors commented on the manuscript.

**Competing interests**
The authors declare no competing interests.

**Figures**

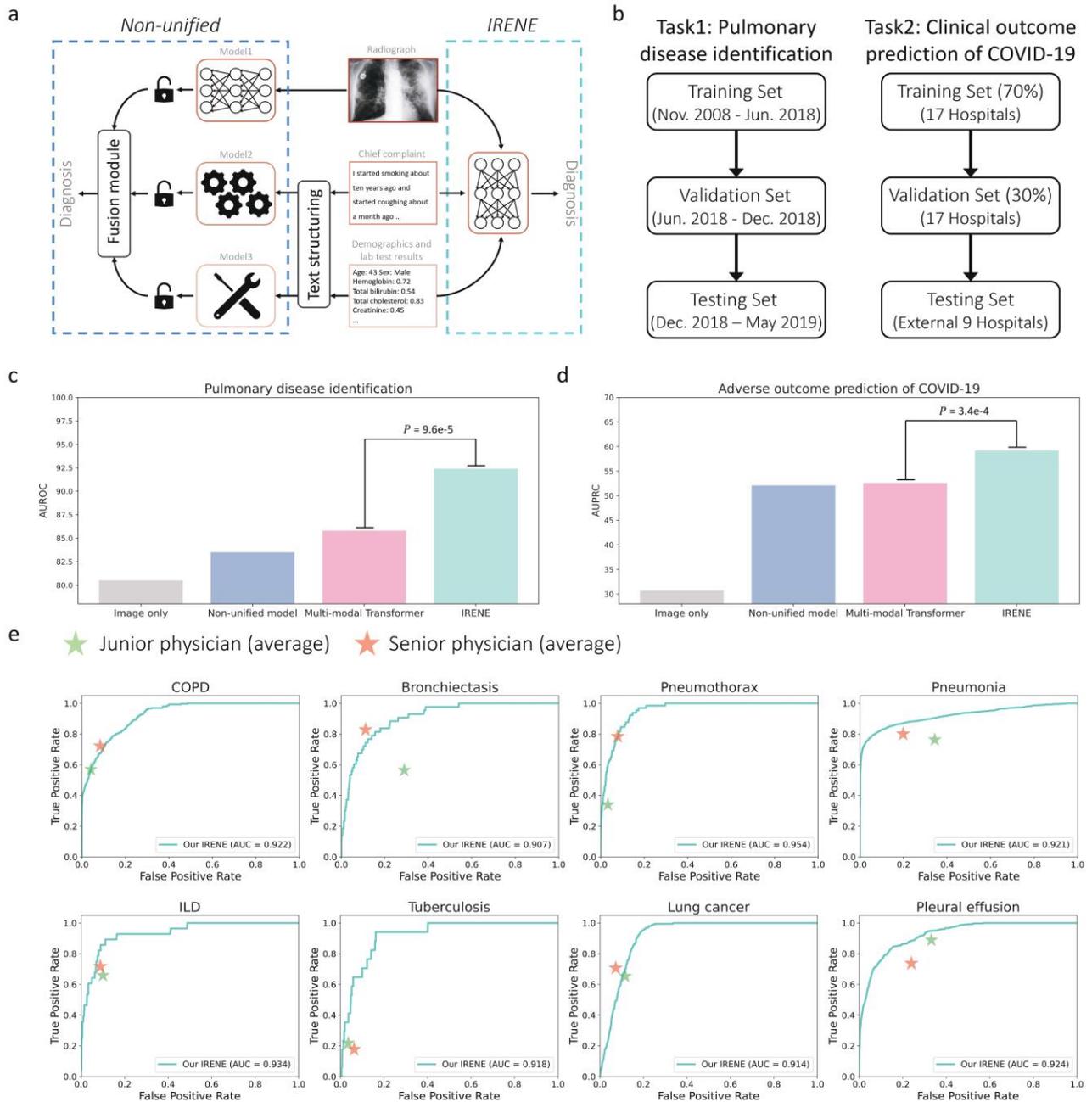

**Fig. 1 | Introduction to IRENE. a** contrasts the previous non-unified multimodal diagnosis paradigm with IRENE. IRENE eliminates the tedious text structuralization process, separate paths for modality-specific feature extraction, and the multimodal feature fusion module in traditional non-unified approaches. Instead, IRENE performs multimodal diagnosis with a single, unified Transformer. **b** shows the scheme for splitting an original dataset into training, validation and testing sets for pulmonary disease identification and adverse clinical outcome prediction of COVID-19, respectively. (**c, d**) compare the experimental results from the image-only models, non-unified early fusion methods, multimodal Transformer (i.e., Perceiver), and IRENE in two tasks. We calculate p-values between the mean performance of IRENE and the multimodal Transformer using the independent two-sample t-test (two-sided). Specifically, we repeat each experiment for ten times with different random seeds, after which p-values are calculated. **e** compares IRENE with junior (with < 7 years of experience) and senior physicians (with more than 7 years of experience). There are two junior physicians and two senior physicians, where average performance within each group is reported. IRENE surpasses the diagnosis performance of junior physicians while performing competitively with senior experts.

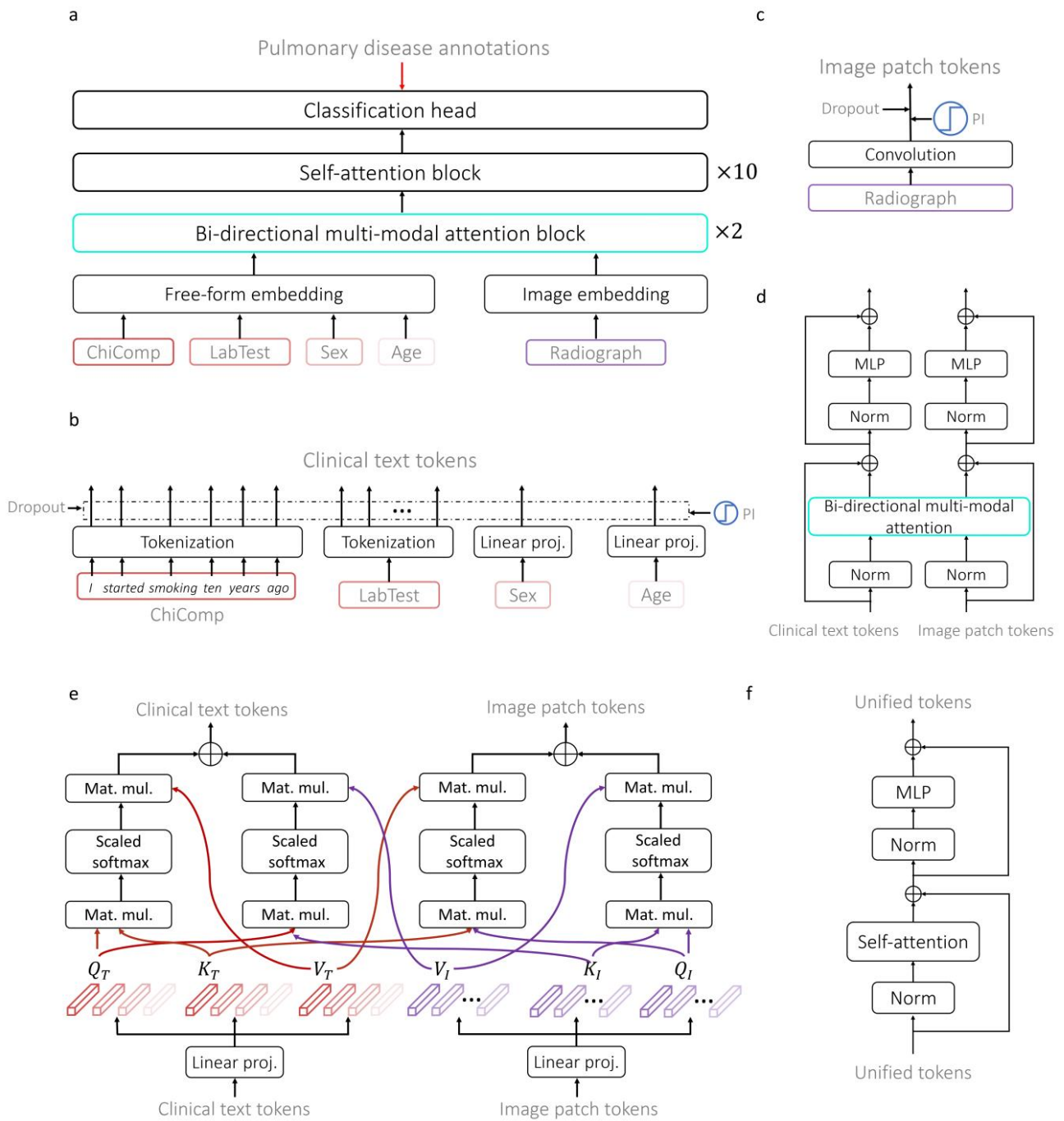

**Fig. 2 | Network Architecture of IRENE. a** shows the overall workflow of IRENE in the first task, i.e., pulmonary disease identification. The input data consist of five parts: the chief complaint (ChiComp), laboratory-test results (LabTest), demographics (Sex and Age), and radiograph. Our multimodal diagnosis Transformer (MDT) includes two bi-directional multimodal attention blocks and ten self-attention blocks. The training process is guided by pulmonary disease annotations provided by human experts. **b** demonstrates how to encode different types of clinical texts in the free-form embedding. Specifically, IRENE accepts unstructured chief complaints as part of the input. **c** shows how to encode a radiograph as a sequence of image patch tokens. **d** presents the detailed design of a bi-directional multimodal attention block, which consists of two layer normalization layers (Norm), one bi-directional multimodal attention layer and one multi-layer perceptron (MLP). **e** presents detailed attention operations in the bi-directional multimodal attention layer, where representations across multiple modalities are learned and fused simultaneously. **f** shows the detailed architecture of a self-attention block.

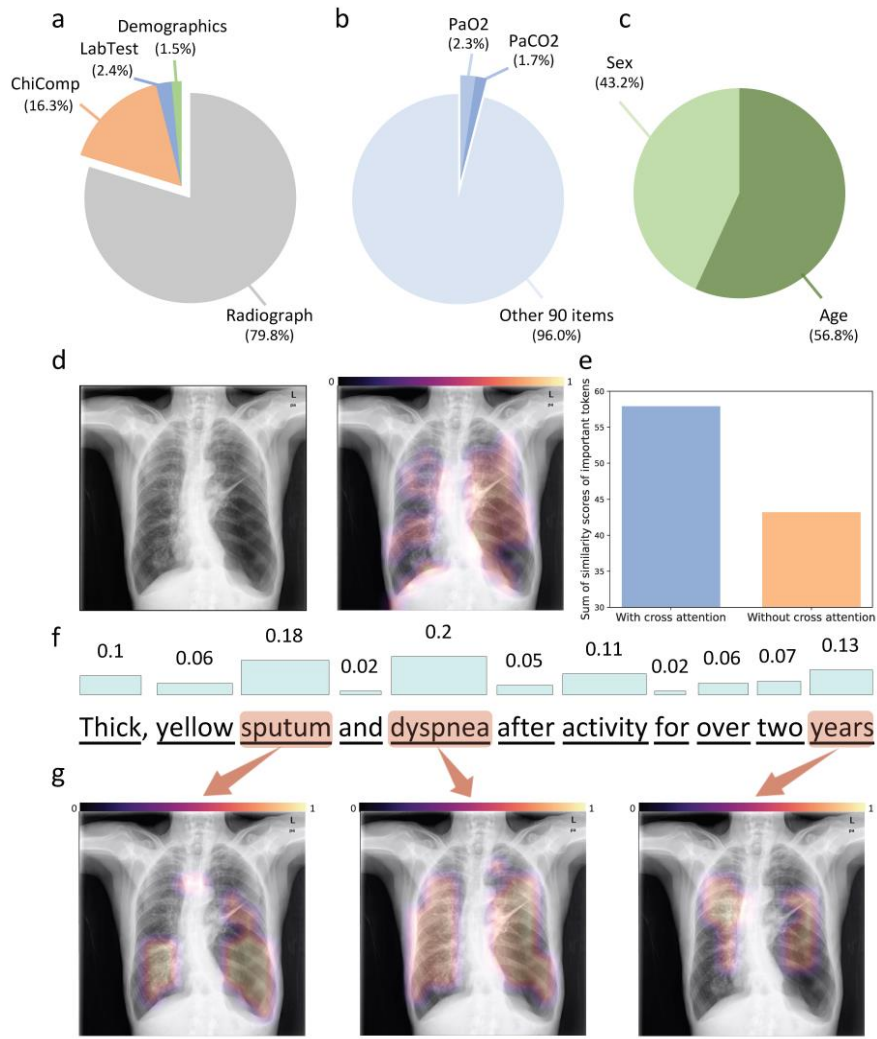

**Fig. 3 | Attention Analysis. a** presents the attention allocated to different types of inputs from a patient with COPD, i.e., the radiograph, chief complaint (ChiComp), laboratory-test results (LabTest), and demographics. **b** shows the relative importance of laboratory-test items. **c** compares the importance of sex and age in making a diagnostic decision. **d** visualizes the attention assigned to individual pixels in the radiograph. The left figure is the input chest X-ray. The right figure presents pixels with different attention values. **e** investigates the impact of cross attention on the relevance and importance of high-ranking words (from chief complaints) and image patches (from radiographs) in the pulmonary disease identification task. Specifically, we define high-ranking words and patches as those whose tokens have top 25% cosine similarity scores with the CLS token. **f** presents the normalized importance of every word in the chief complaint. **g** visualizes the distribution of attention between every image patch and each of the top 3 ranked words. The color bar in (**d**, **g**) illustrates IRENE's confidence about a pixel being abnormal, where a bright color stands for high confidence, and a dark color denotes low confidence.

**Table 1 | Comparison with baseline models in the task of pulmonary disease identification.** The baseline models include the image-only model, the early fusion method, the late fusion approach, and two recent Transformer-based multimodal classification models (i.e., GIT and Perceiver). 95% CI denotes the 95% confidence interval. The evaluation metric is AUROC.

| Method | Mean | COPD | Bronchiectasis | Pneumothorax | Pneumonia | ILD | Tuberculosis | Lung cancer | Pleural effusion |
|---|---|---|---|---|---|---|---|---|---|
| Image-only | 0.805 (0.802, 0.808) | 0.847 (0.845, 0.851) | 0.746 (0.743, 0.748) | 0.789 (0.786, 0.791) | 0.845 (0.843, 0.848) | 0.799 (0.796, 0.801) | 0.769 (0.765, 0.772) | 0.825 (0.821, 0.830) | 0.819 (0.817, 0.822) |
| Early Fusion | 0.835 (0.832, 0.839) | 0.895 (0.893, 0.898) | 0.772 (0.768, 0.775) | 0.810 (0.807, 0.812) | 0.873 (0.870, 0.875) | 0.824 (0.822, 0.827) | 0.793 (0.791, 0.796) | 0.871 (0.868, 0.875) | 0.842 (0.839, 0.845) |
| Late Fusion | 0.826 (0.823, 0.828) | 0.888 (0.885, 0.890) | 0.765 (0.763, 0.767) | 0.822 (0.820, 0.825) | 0.870 (0.868, 0.872) | 0.804 (0.802, 0.805) | 0.770 (0.767, 0.772) | 0.839 (0.836, 0.841) | 0.850 (0.847, 0.852) |
| GIT | 0.848 (0.844, 0.850) | 0.911 (0.907, 0.913) | 0.798 (0.796, 0.800) | 0.824 (0.821, 0.827) | 0.895 (0.893, 0.898) | 0.819 (0.816, 0.821) | 0.807 (0.804, 0.810) | 0.872 (0.871, 0.873) | 0.858 (0.855, 0.860) |
| Perceiver | 0.858 (0.855, 0.861) | 0.910 (0.907, 0.912) | 0.788 (0.784, 0.791) | 0.846 (0.842, 0.850) | 0.903 (0.901, 0.906) | 0.830 (0.827, 0.833) | 0.825 (0.823, 0.828) | 0.890 (0.887, 0.892) | 0.872 (0.869, 0.874) |
| IRENE | 0.924 (0.921, 0.926) | 0.922 (0.920, 0.925) | 0.907 (0.903, 0.910) | 0.954 (0.952, 0.957) | 0.921 (0.918, 0.923) | 0.934 (0.929, 0.937) | 0.918 (0.917, 0.921) | 0.914 (0.911, 0.917) | 0.924 (0.921, 0.926) |

**Table 2 | Comparison with baseline models in the task of adverse clinical outcome prediction of COVID-19 patients.** We included five models in the comparison, which are the image-only model, the early fusion method, the late fusion approach, and two recent Transformer-based multimodal classification models (i.e., GIT and Perceiver). 95% CI denotes the 95% confidence interval. The evaluation metric is AUPRC.

| Method | Mean | Admission to ICU | Need for MV | Death |
|---|---|---|---|---|
| Image-only | 0.307 (0.237, 0.391) | 0.482 (0.355, 0.636) | 0.247 (0.136, 0.398) | 0.192 (0.073, 0.333) |
| Early Fusion | 0.521 (0.435, 0.614) | 0.665 (0.548, 0.774) | 0.551 (0.397, 0.699) | 0.346 (0.174, 0.544) |
| Late Fusion | 0.503 (0.422, 0.598) | 0.647 (0.535, 0.759) | 0.533 (0.388, 0.685) | 0.330 (0.164, 0.531) |
| GIT | 0.514 (0.442, 0.605) | 0.653 (0.546, 0.743) | 0.554 (0.411, 0.702) | 0.335 (0.168, 0.554) |
| Perceiver | 0.526 (0.448, 0.611) | 0.652 (0.529, 0.771) | 0.566 (0.406, 0.715) | 0.360 (0.201, 0.543) |
| IRENE | 0.592 (0.500, 0.682) | 0.712 (0.587, 0.834) | 0.624 (0.473, 0.754) | 0.441 (0.270, 0.617) |

**Table 3 | An ablation study of IRENE by removing or replacing individual components.** HA (N) denotes the presence of N bi-directional multimodal attention block(s) in the multimodal diagnosis Transformer (MDT) while the remaining blocks are self-attention blocks (twelve blocks in total). Image denotes the input radiograph. Uni-direction means we only compute text-to-image attention in multimodal attention blocks. ChiComp stands for the chief complaint. LabTest denotes laboratory-test results. Tokenization stands for the tokenization procedures for the chief complaint and laboratory-test results. The evaluation metric is AUROC.

| Row | HA (2) | HA (0) | HA (6) | Uni-direction | Image | ChiComp | LabTest | Tokenization | Mean |
|---|---|---|---|---|---|---|---|---|---|
| 0 | √ |   |   |   |   | √ | √ | √ | 0.924 (0.921, 0.926) |
| 1 |   | √ |   |   | √ | √ | √ | √ | 0.858 (0.850, 0.867) |
| 2 |   |   | √ |   | √ | √ | √ | √ | 0.905 (0.899, 0.910) |
| 3 | √ |   |   | √ | √ | √ | √ | √ | 0.884 (0.880, 0.888) |
| 4 | √ |   |   |   | √ |   | √ | √ | 0.860 (0.855, 0.864) |
| 5 | √ |   |   |   | √ | √ |   | √ | 0.882 (0.873, 0.891) |
| 6 | √ |   |   |   | √ | √ | √ |   | 0.894 (0.886, 0.900) |
| 7 | √ |   |   |   |   | √ | √ | √ | 0.543 (0.525, 0.569) |